\documentclass[10pt,twocolumn,letterpaper]{article}

\usepackage{iccv}
\usepackage{times}
\usepackage{epsfig}
\usepackage{graphicx}
\usepackage{amsmath}
\usepackage{amssymb}
\usepackage[colorlinks,linkcolor=blue]{hyperref}

\iccvfinalcopy % *** Uncomment this line for the final submission

\def\iccvPaperID{****} % *** Enter the ICCV Paper ID here

\ificcvfinal\fi

\begin{document}

%%%%%%%%% TITLE
\title{Top-1 Solution of  Multi-Moments in Time Challenge 2019 }

\renewcommand{\thefootnote}{}
\newcommand*{\affaddr}[1]{#1} % No op here. Customize it for different styles.
\newcommand*{\affmark}[1][*]{\textsuperscript{#1}}
\newcommand*{\email}[1]{\texttt{#1}}
\author{%
Manyuan Zhang\affmark[1*]  \ \ \ \ Hao Shao\affmark[1*] \ \ \ \ Guanglu Song\affmark[1]\ \ \ \ Yu Liu \affmark[2]\ \ \ \ Junjie Yan\affmark[1]\\
\affaddr{\affmark[1]SenseTime X-Lab}\\
\affaddr{\affmark[2]Multimedia Laboratory, The Chinese University of Hong Kong}\\
{\small \email{\ \ {\{zhangmanyuan,shaohao\}@sensetime.com}}
}}

\maketitle
\ificcvfinal\thispagestyle{empty}\fi

%%%%%%%%% ABSTRACT
\begin{abstract}
In this technical report, we briefly introduce the solutions of our team 'Efficient' for the Multi-Moments in Time challenge in ICCV 2019. We first conduct several experiments with popular Image-Based action recognition methods TRN, TSN, and TSM. Then a novel temporal interlacing network is proposed towards fast and accurate recognition. Besides, the SlowFast network and its variants are explored. Finally, we ensemble all the above models and achieve 67.22\% on the validation set and 60.77\% on the test set, which ranks 1st on the final leaderboard. In addition, we release a new code repository for video understanding which unifies state-of-the-art 2D and 3D methods based on PyTorch. The solution of the challenge is also included in the repository, which is available at \url{ https://github.com/Sense-X/X-Temporal}.

\end{abstract}

%%%%%%%%% BODY TEXT

\section{Introduction of the Challenge}

The Moments in Time Multimodal Multi-Label Action Detection Challenge at ICCV'19 is as part of the Workshop on Multi-modal Video Analysis. The challenge requires the algorithm to detect the event labels depicted in trimmed videos. It contains over one million training and testing videos, and each video has multiple action labels, which is currently the largest multi-label video classification challenge.

\footnote{*They contributed equally to this work.}
\section{Overview of  Methods}
In this work, we have experimented with both image-based and 3D-based methods. Image-based  methods includes  TSN\cite{wang2016temporal}, TSM \cite{lin2018temporal}, TRN \cite{zhou2018temporal} and TIN\cite{shao2020temporal}. These methods all use 2D convolution kernels instead of 3D convolution kernels to capture the temporal information. The number of their parameters and FLOPs are small compared to 3D-Based Models, but most of them don't have better performance than those 3D Networks. For the 3D-based methods, we conduct experiments mainly based on the SlowFast\cite{feichtenhofer2018slowfast} and several of its efficient variants. In following sections, we will first introduce  the image-based models, then are the 3D-based methods. All the experiment results and conclusions are given subsequently.

\section{Image-Based Models}

%-------------------------------------------------------------------------
\subsection{Temporal Segment Network}
Temporal segment network (TSN~\cite{wang2016temporal}) is a framework for video-based action recognition. TSN takes a strategy of sampling a fixed number of sparse segments from one video to model long-term temporal structure. The final prediction of video is averaged by the logits of each chip. In this competition, we experimented with the temporal segment network with evenly sampling 5 segments form one video.

\subsection{Temporal Relational Network}
Temporal relational network (TRN~\cite{zhou2018temporal}) is a recognition framework that can model and reason about temporal dependencies between different segments of one video. The model is also designed to reason at multiple time scales. However, it doesn't work well in our attempt.

\subsection{Temporal Shift Module}
Temporal Shift Module (TSM~\cite{lin2018temporal}) proposes an operator that shifts part of the channels along the temporal dimension. The operator can help the network fuse the temporal information among neighboring frames. We experimented the model with different backbones and input sequence lengths T.

\subsection{Temporal Interlacing Network}
\begin{figure}[htb]
\centering
\includegraphics[width=1.0\linewidth]{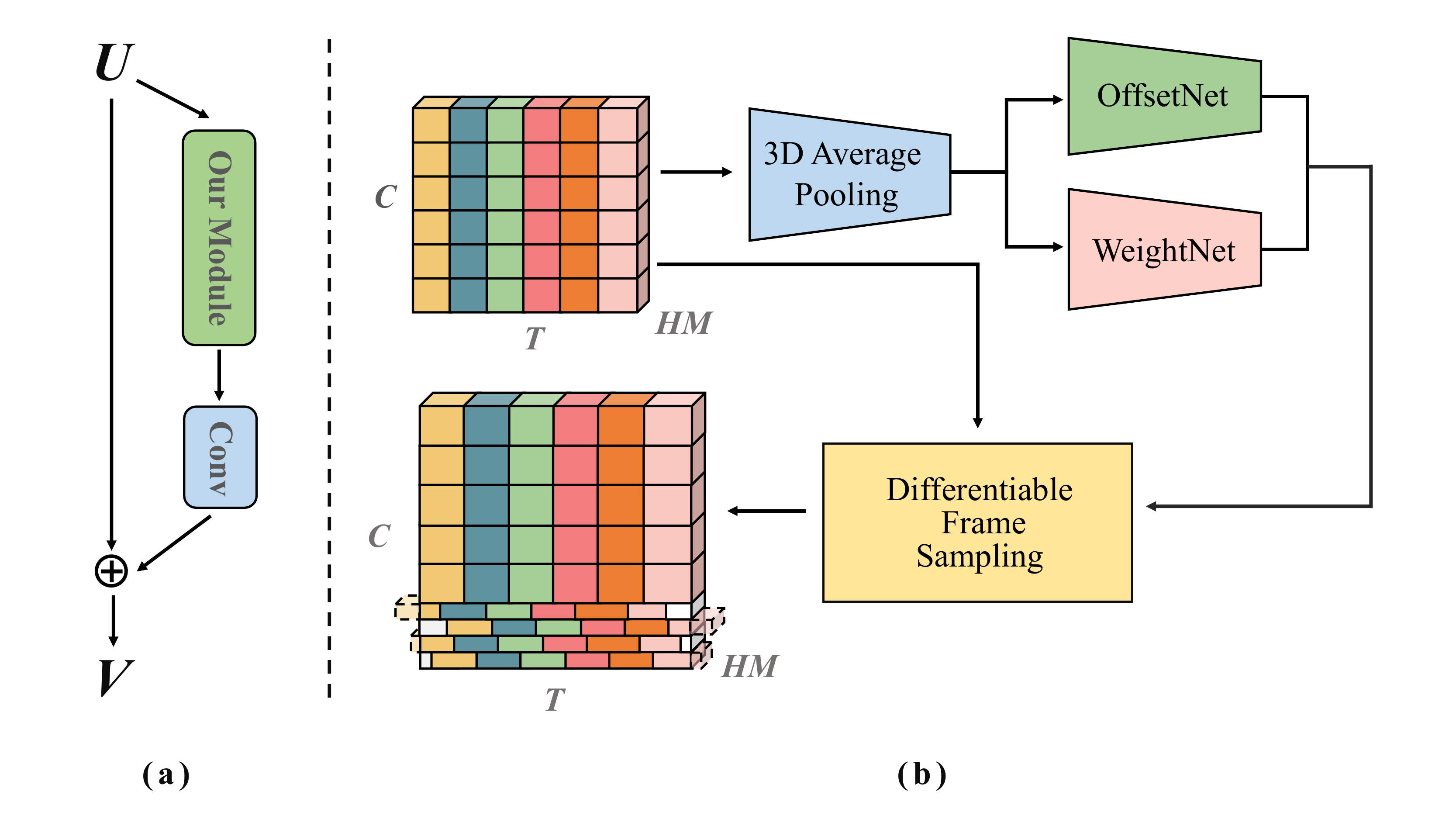}
\caption{Architecture of TIN: our input video clips composed of uniformly sampled from raw frames are processed by our modified 2D ResNet-50. \textbf{(a)} We plug our module before the convolution layer in each block. \textbf{(b)} Our module obtains offsets and weights by OffsetNet and WeightNet, then shifts and samples the feature along with the temporal dimension.}
\label{fig:archi}
\end{figure}

\begin{figure}[htb]
\centering
\includegraphics[width=1.0\linewidth]{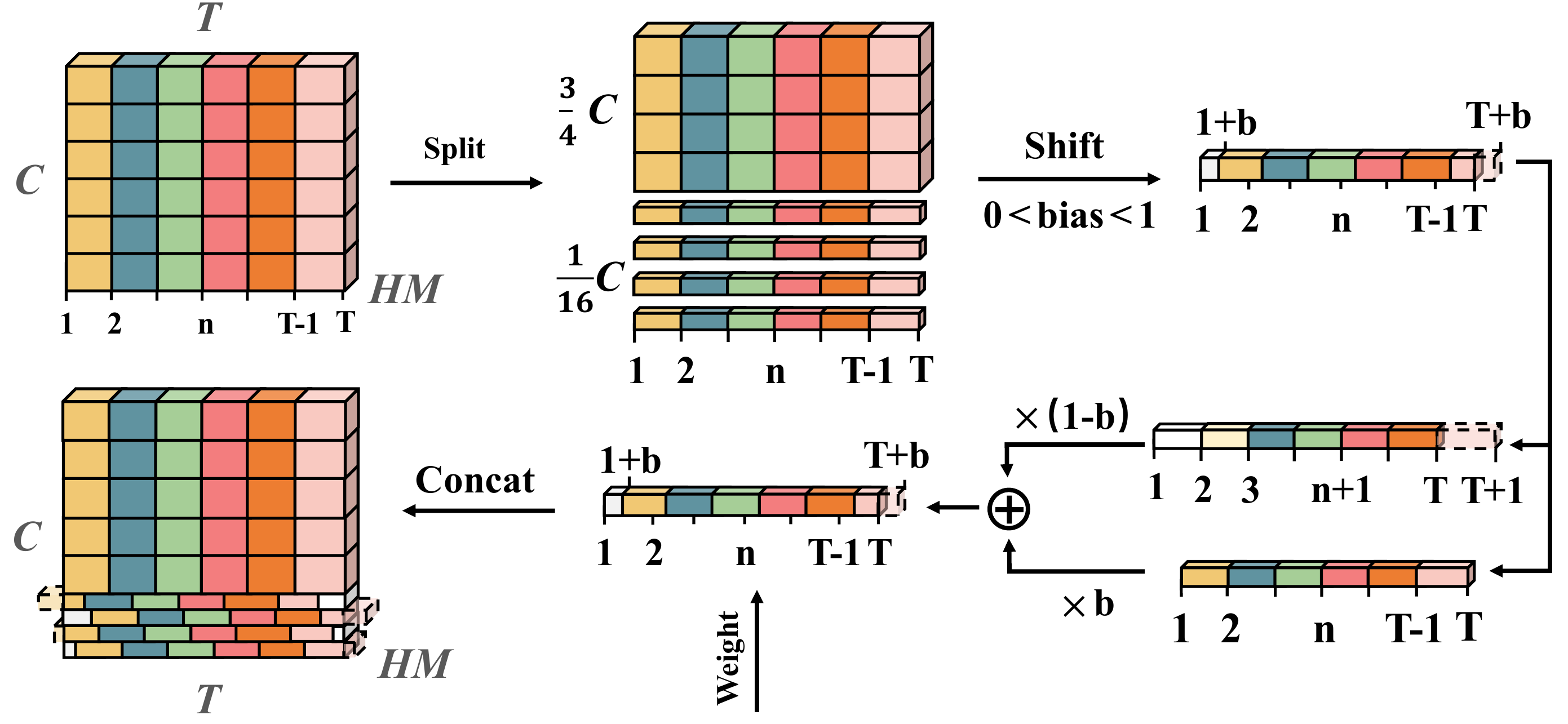}
\caption{The pipeline of differentiable Temporal-wise Frame Sampling. The module splits the feature map into several groups and shifts them by different offsets according to their group. Then compute the weighted sum along with the temporal dimension. Finally, we concatenate the split groups into the integral feature map, which is the same size as the input data.}
\label{fig:sample}
\end{figure}

In this work, we proposed a Temporal Interlacing Network (TIN~\cite{shao2020temporal}) which uses a network to model the relation between the shift distance and the specific input data. We fuse temporal information along the temporal dimension through inserting our module before each convolutional layer in the residual block, as illustrated in Figure.~\ref{fig:archi} and Figure.~\ref{fig:sample}. While TSM can only shift the channels along the temporal dimension by +1 or -1. The differentiable module we designed can infer the suitable displacement length according to different groups and suitable weight for the feature-map along the temporal dimension. Our proposed module has almost the same FLOPs and parameters as the origin TSM model. Moreover, in our experiments between TSM and TIN, TIN obtained about 1\% - 2\% better performance with the same train and test configure.

\section{SlowFast-based Models}

In this part, we conduct experiments on SlowFast~\cite{feichtenhofer2018slowfast} network. SlowFast network has two paths, a slow path to capture appearance content while a fast path to capture motion information. For details about the architecture, please refer to its raw publication~\cite{feichtenhofer2018slowfast}.

For this challenge, we train several SlowFast models and its variants. Note that only RGB input is used, for the reason that flow extraction costs too much computation and storage. The models we select are (a) SlowFast, Slow path $8\times8$ and Fast path  $32\times2$, with an input clip of consecutive 64 frames (b) only Fast path $32\times2$, with no channel reduction. This model is pretty heavy. It has above 4x computation consumption than the standard SlowFast network. (c) only Slow path $8\times8$, which only keeps the slow path to capture appearance content. (d) SlowFast, Slow path $11\times8$ and Fast path  $44\times2$. Due to most videos has around 90 frames, this model designs to capture the whole video information.

\begin{table*}[t]
\centering
\begin{tabular}{cccc}
\hline
 & mAP@clip & \begin{tabular}[c]{@{}c@{}}mAP@video\\ (dense testing)\end{tabular} & \begin{tabular}[c]{@{}c@{}}mAP@video\\ ( multi-scale \& dense testing)\end{tabular} \\ \hline
SlowFast, 8$\times$8 & 59.89 & 61.52 & 63.52 \\ 
SlowFast, 11$\times$8 & 60.41 & 61.60 & 64.00 \\ 
Fast,16$\times$2 & 59.05 & 61.55 & 63.45 \\ 
Fast, 32$\times$2 & 62.03 & 63.59 & \textbf{64.81} \\ 
Slow, 8$\times$4 & 59.02 & 61.68 & 62.99 \\ \hline
Ensemble & \multicolumn{2}{c}{66.71~(val)} & 59.80~(test) \\ \hline
Ensemble+Image-based methods & \multicolumn{2}{c}{\textbf{67.22}~(val)} & \textbf{60.77}~(test) \\ \hline
\end{tabular}
\caption{Results for SlowFast-based networks on the validation set. By ensemble with Image-based methods, we achieve 67.22\% on validation set and 60.77\% on test set, which ranks 1st on the final leaderboard. }
\label{tab:3d map}
\end{table*}

\begin{table}[t]
\centering
\caption{mAP on the validation set of the Multi Moments in Time Dataset.}
\begin{tabular}{llll}
\hline
\textbf{Model} & \textbf{Frames} & \textbf{Backbone} & \textbf{mAP} \\ \hline
TSM            & 8               & ResNet-50         & 59.85        \\
TSM            & 16              & ResNet-50         & 61.12        \\
TSM            & 8               & ResNet-101        & 61.06        \\ \hline
TIN           & 8               & ResNet-50         & 62.23        \\
TIN           & 8               & ResNet-101        & 62.22        \\
TIN           & 16              & ResNet-101        & 62.49        \\ \hline
TSN            & 5               & ResNet-101        & 58.92        \\
TSN            & 5               & IC-ResNet-v2      & 57.45        \\
TSN            & 5               & SeNet-154         & 53.19        \\ \hline
\end{tabular}

\label{map on 2D}
\end{table}
\textbf{Training}. For the training of Image-Based Models, we evenly select 8 or 16 frames from a video and the spatial size of the input image is 224$\times$224 pixels. About the data augmentation, we use the technique of spatial jittering and horizontal flipping to alleviate the problem of overfitting. We set the dropout rate~\cite{srivastava2014dropout} to 0.5 and set weight decay to 5e-4. Meantime, we use the algorithm of mini-batch stochastic gradient descent with a momentum of 0.9. The initial learning rate is set to 0.01 and divided by 10, 20 epochs.

\section{Experiments}

\textbf{Dataset.} The Multi-Moments in Time datasets ~\cite{monfort2019moments} consists of more than one million three-second videos with two million action labels. This dataset focus on multi-label action recognition in videos. There are 1,025,862 training ,~10,000 validation and 10,000 test videos segments. We use FFmpeg to extract frames with 30 fps and resize the short side to 320 pixels. Note that 737 training videos, 8 validation videos, and 6 test videos can't extract frames. The performance metric is mean Average Precision (mAP) under sample level.

\textbf{Loss function.} For the multi-label action recognition problem, there exist  high relationships between class labels. Proper loss function will capture the relationship and leading to better convergence. We try several multi-label classification loss functions including Warp~\cite{li2017improving},  Lsep~\cite{weston2011wsabie}, and most used binary cross-entropy loss.  During the experiments, we find that simply scale up BCE loss  achieves a huge performance gain. But scale up learning rate with the same ratio can not have the same performance. Scale-up BCE loss which uses 'mean' as a reduction operator in PyTorch with factor 160 achieves the best performance and we use it as our loss function.

\textbf{Class unbalance.} we count the sample number for each class, the max number of videos in one class is 48060 while the min is 504. To tackle the training sample unbalance problem, we give each class a loss weight. Denotes the mean number of videos in a class as $N_{mean}$, number videos in class $i$ as $N_{i}$. We have two strategies, weighted by $\sqrt{\frac{N_{i}}{N_{mean}}}$ and $\frac{N_{i}}{N_{mean}}$. Both of them achieve slightly performance gain, so we directly use original data to train, without handling the class sample unbalance problem.

\textbf{Model pre-train.} There are several works claims that pre-train is important for action recognition~\cite{carreira2017quo,ghadiyaram2019large}. We pre-train  models on Kinetics dataset, which is the largest scale well-label action recognition dataset. Pre-train improves Image-based methods but it is interesting that pre-train on Kinetics hurt recognition performance for SlowFast-based methods (1.1\% drop). The reason may be that Kinetics focuses on human activities but Multi-Moments In Time dataset includes a wider range of actions like animal and nature motion.

For training the SlowFast-based model, we use the half-period cosine learning rate ~\cite{loshchilov2016sgdr} for training. Learning rate at $k$-th iteration is $\eta \cdot 0.5\left[\cos \left(\frac{\dot{n}}{n_{\max }} \pi\right)+1\right]$, the base learning rate is 0.2 and the max iteration number is 180k. We also adopt linear warm up~\cite{goyal2017accurate} for stable convergence. The momentum is 0.9 and weight decay is $10^{-4}$. Dropout ~\cite{hinton2012improving}  is set to 0.5 to prevent over-fitting. 32 Nvidia V100 GPUs with synchronized SGD and synchronized BatchNorm is used. The mini-batch size in one GPU is 32,  1024 in total. The training process usually lasts for a week. As for data augmentation, we first random sample video clips with consecutive frames. For the spatial domain, we randomly resize the video frame short side to  [128,160] and then randomly crop $112\times112$ pixels. Randomly  flip is also  used.

\subsection{Results for Image-Based Models}

In this competition, we use ResNet-50~\cite{he2016deep}, ResNet-101, Inception-ResNet-v2, Senet-151 as our backbone models, which are pretrained on Kinetics-600~\cite{carreira2018short}. The mAP on the validation set of the Multi Moments in Time are shown in Table.~\ref{map on 2D} We can find that models with more input frames have better performance, and TIN achieved 2.37\% better performance compared to TSM with the same 8 frames and backbone network. From Table.~\ref{map on 2D}, we can also find that TSN still obtained a not bad performance with ResNet-101 backbone networks.

We also try to sample more chips from one video, it can also help the model get higher mAP on the validation set. In detail, we sample 5 random chips along the temporal dimension. With respect to the spatial dimension, we also random crop 3 chips which the size of it is 256$\times$256 pixels. These tricks can help our models achieve about 0.5\%-1.1\% improvement compared with using only one chip to test.

\subsection{Results for SlowFast-Based Methods}

In this competition, we select SlowFast-101 as base model. Deeper network SlowFast-152 and SlowFast-200 is also explored, but they achieve near the same performance as SlowFast-101, so we finally conduct experiments base on SlowFast-101. The results for SlowFast network and its variants are shown in Table.~\ref{tab:3d map}.  From Table.~\ref{tab:3d map}, we can find that only Fast path with 64 consecutive frames input achieves the best performance, due to its huge computation. 

We also conduct multi-scale and multi-crop inference.  For each test video, we uniformly sample 10 clips along its temporal axis. For each clip, we scale the shorter size to 128 and random crop 3 clips with $128\times128$ pixels from the spatial domain. A multi-scale crop like 144 and 160 is also applied for the robust test. For all crop clips from a video, class prediction probabilities are average. Results are list in the Table.~\ref{tab:3d map}. The multi-scale and dense inference strategy leading to around 2\% performance gain.

\section{Conclusion}

In our submission to the Multi-Moments in Time Challenge 2019, we experiment with many popular video action methods including TSN, TRN, TSM, and SlowFast. A novel Image-based recognition model TIN is proposed to achieve fast and accuracy trade-off.  For this work, only RGB input is used, multimodal input including flow and audio will be explored in the further work. A more sophisticated loss function can also be studied for  multi-label action recognition problem.

{\small
\bibliographystyle{ieee_fullname}
\bibliography{egbib}
}

\end{document}